\documentclass[sigconf]{acmart}

\AtBeginDocument{%
  \providecommand\BibTeX{{%
    \normalfont B\kern-0.5em{\scshape i\kern-0.25em b}\kern-0.8em\TeX}}}

\copyrightyear{2021} 
\acmYear{2021} 
\setcopyright{acmcopyright}
\acmConference[MM '21]{Proceedings of the 29th ACM International Conference on Multimedia}{October 20--24, 2021}{Virtual Event, China}
\acmBooktitle{Proceedings of the 29th ACM International Conference on Multimedia (MM '21), October 20--24, 2021, Virtual Event, China}
\acmPrice{15.00}
\acmDOI{10.1145/3474085.3475303}
\acmISBN{978-1-4503-8651-7/21/10}

\acmSubmissionID{mfp0778}

\usepackage{amsmath}
\usepackage{multirow}
\usepackage{booktabs}
\usepackage{enumitem}
\usepackage{overpic}

\newcommand{\upoc}[1]{{\emph{$\text{UPOC}^\text{2}$}}}

\renewcommand\footnotetextcopyrightpermission[1]{} % removes footnote with conference information in first column
\begin{document}
\fancyhead{}

\title{Product-oriented Machine Translation with Cross-modal Cross-lingual Pre-training}

\author{Yuqing Song}
\affiliation{%
 \institution{Renmin University of China}
 \country{}
}
\email{syuqing@ruc.edu.cn}

\author{Shizhe Chen}
% \authornote{This work was performed when Shizhe Chen was at Renmin University of China.}
\affiliation{%
 \institution{INRIA}
 \country{}
}
\email{shizhe.chen@inria.fr}

\author{Qin Jin}
\authornote{Corresponding Author}
\affiliation{%
 \institution{Renmin University of China}
 \country{}
}
\email{qjin@ruc.edu.cn}

\author{Wei Luo}
\affiliation{%
 \institution{Alibaba Damo Academy}
 \country{}
}
\email{muzhuo.lw@alibaba-inc.com}

\author{Jun Xie}
\affiliation{%
 \institution{Alibaba Damo Academy}
 \country{}
}
\email{qingjing.xj@alibaba-inc.com}

\author{Fei Huang}
\affiliation{%
 \institution{Alibaba Damo Academy}
 \country{}
}
\email{f.huang@alibaba-inc.com}


\begin{abstract}
Translating e-commercial product descriptions, a.k.a product-oriented machine translation (PMT), is essential to serve e-shoppers all over the world.
However, due to the domain specialty, the PMT task is more challenging than traditional machine translation problems.
Firstly, there are many specialized jargons in the product description, which are ambiguous to translate without the product image.
Secondly, product descriptions are related to the image in more complicated ways than standard image descriptions, involving various visual aspects such as objects, shapes, colors or even subjective styles.
Moreover, existing PMT datasets are small in scale to support the research.
In this paper, we first construct a large-scale bilingual product description dataset called Fashion-MMT, which contains over 114k noisy and 40k manually cleaned description translations with multiple product images.
To effectively learn semantic alignments among product images and bilingual texts in translation, we design a unified product-oriented cross-modal cross-lingual model (\upoc~) for pre-training and fine-tuning.
Experiments on the Fashion-MMT and Multi30k datasets show that our model significantly outperforms the state-of-the-art models even pre-trained on the same dataset.
It is also shown to benefit more from large-scale noisy data to improve the translation quality.
We will release the dataset and codes at \url{https://github.com/syuqings/Fashion-MMT}.
\end{abstract}

\begin{CCSXML}
<ccs2012>
<concept>
<concept_id>10010147.10010178.10010179.10010180</concept_id>
<concept_desc>Computing methodologies~Machine translation</concept_desc>
<concept_significance>500</concept_significance>
</concept>
<concept>
<concept_id>10010147.10010178.10010179.10010182</concept_id>
<concept_desc>Computing methodologies~Natural language generation</concept_desc>
<concept_significance>500</concept_significance>
</concept>
</ccs2012>
\end{CCSXML}

\ccsdesc[500]{Computing methodologies~Machine translation}
\ccsdesc[500]{Computing methodologies~Natural language generation}

\keywords{Product-oriented Machine Translation; Pre-training; Dataset; Multimodal Transformer}

\maketitle

\vspace{3pt}
\section{Introduction}
With the rapid development of e-commerce, more and more people go shopping online because of its convenience and efficiency.
In order to better serve e-shoppers all over the world, it is necessary to translate e-commercial product descriptions into various languages.
Therefore, the product-oriented machine translation (PMT) task \cite{zhou2018ikea, calixto2017ecommerce, calixto2017casestudy} has received growing research attentions recently.

\begin{figure}[t]
  \centering
  \begin{overpic}[scale=0.33]{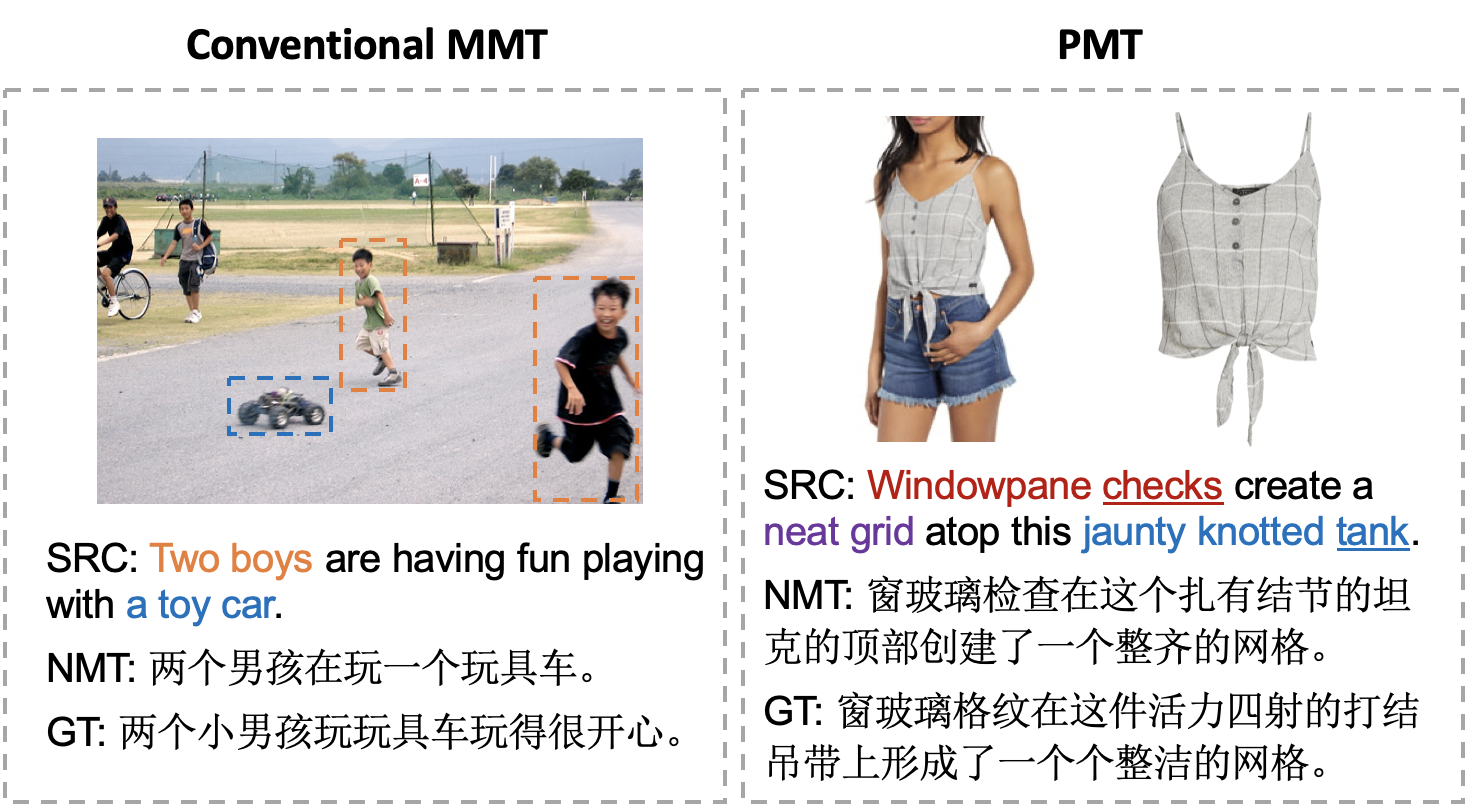}
  \end{overpic}
  \caption{Examples of MMT and PMT tasks. The colored text are visually relevant phrases. The underlined words are specialized jargons which are ambiguous and translated incorrectly by current machine translation system.}
  %\vspace{-10pt}
  \label{fig:intro}
\end{figure}

The domain specialty makes the PMT task more challenging than traditional machine translation problems.
Firstly,  product descriptions contain many specialized jargons, which could be ambiguous in different contexts.
It is hard to understand and translate these descriptions without the corresponding product images.
For example, in Figure~\ref{fig:intro}, the word ``checks'' in the source product description means ``grids'' and the word ``tank'' means ``vest''. The meanings of these two words are different from their common meanings. Therefore, the current text-based machine translation system cannot translate them correctly. 
Secondly, although the visual context is beneficial to the translation, the relevance between product image and text description is more complex than that in conventional multimodal machine translation (MMT) task.
As shown in Figure~\ref{fig:intro}, in the MMT task, the text description explicitly describes the major objects in the image. 
However, in the PMT task, text descriptions are related to images in very different aspects, such as products, shapes, colors or even more subjective styles.
Therefore, it requires PMT models to dynamically extract different types of information from images to help the text translation.
Last but not least, existing resources for PMT research are rather limited, for example, the latest PMT dataset IKEA \cite{zhou2018ikea} contains only 3,600 data samples in the <image, source sentence, target translation> triplet format.
It is extremely expensive to manually annotate translations for product descriptions as it demands the annotator to master multiple languages as well as knowledge about the products. 
The scarcity of PMT datasets further constrains the automatic translation quality of product descriptions.

In this paper, we construct a large-scale bilingual product description dataset, Fashion-MMT, which is based on automatically crawled product images and English descriptions from e-commerce website \cite{yang2020facad}.
The dataset covers various fashion products from 78 categories, including dresses, shoes, pants, sunglasses, earrings and so on.
We create two types of translation annotations. The first type, called Fashion-MMT(L), contains 114,257 automatic Chinese translations of original English product descriptions via a state-of-the-art text-based machine translation system.
Although it is easy to achieve large-scale, such translations are noisy.
%It can be obtained easily in large-scale but such translations are noisy.
The second type, called Fashion-MMT(C), is a cleaned subset of Fashion-MMT(L) and contains 40,000 <image, English description, Chinese translation> triplets with the manually annotated Chinese translations.

In order to take advantage of the large-scale Fashion-MMT dataset to learn semantic alignments among product images and bilingual texts, we propose a Unified Product-Oriented Cross-lingual Cross-modal (\upoc~) model for pre-training and fine-tuning.
The \upoc~~model is based on multimodal transformer \cite{lu2019vilbert, li2019visualbert, chen2019uniter} and we design three proxy tasks to effectively learn the inter relationships among images and bilingual texts in pre-training, including multi-modal translation language modeling (MTLM), image source-sentence matching (ISM), and product attribute prediction (ATTP) tasks.
Experimental results on Fashion-MMT(C) and Multi30k benchmark datasets show that even pre-trained on the same dataset without any extra data, our model outperforms the state-of-the-art model by a large margin due to its better abilities in cross-modal fusion, i.e., +5.9\% BLEU@4 score on Multi30k and +2.1\% on Fashion-MMT(C).
When augmented with large-scale noisy triplet data in Fashion-MMT(L), our model achieves more gains by +1.4 BLEU@4 score and +17.6 CIDEr score.

The main contributions of our work are summarized as follows:
\parskip=0.1em
\begin{itemize}[itemsep=1pt,partopsep=0pt,parsep=\parskip,topsep=0.5pt]
    \item We propose the first large-scale PMT dataset Fashion-MMT to support the PMT research and e-commerce applications.
    \item We design a unified pre-training and fine-tuning model \upoc~ and three cross-modal cross-lingual proxy tasks to enhance product machine translation.
    \item Our model achieves the state-of-the-art results on both the Multi30k and Fashion-MMT datasets, and demonstrates its robustness to benefit from large-scale noisy data.
\end{itemize}

\vspace{2pt}
\section{Related Works}
\noindent\textbf{Multimodal Machine Translation.}
The common visual world is shown to be beneficial to bridge different languages \cite{chen:bilingual,sigurdsson:visual_grounding,chen2019fromwordstosents,huang2020ummt,su2019ummt}.
Therefore, recent research works have paid more attentions to the Multimodal Machine Translation (MMT) task \cite{elliott2016Multi30k}, which aims to leverage visual context to aid textual machine translation.
Calixto and Liu \cite{calixto-liu:Init} explore to integrate global image feature into the encoder or decoder of the NMT model, while works in \cite{caglayan:AttConcat, calixto:doublyAtt, helcl:attentionStrategy, delbrouck:gating, yao2020mmt} employ more fine-grained image context such as spatial image regions via attention mechanism.
Caglayan et al. \cite{caglayan:AttConcat} and Calixto et al. \cite{calixto:doublyAtt} apply concatenation or mean pooling on two independently attended contexts, including the textual context of source words and visual context of image regions, when predicting each target word. 
% Libovick{\'{y}} and Helcl \cite{helcl:attentionStrategy} propose a hierarchical attention strategy to combine the two attended contexts more effectively.
To avoid noise of irrelevant information in the image,  Delbrouck and Dupont \cite{delbrouck:gating} propose a gating mechanism to weight the importance of the textual and image contexts.
Recently, more works \cite{Ive2019distilling, yang2020visual_agreement, yin2020graph, lin2020capsule} propose to represent the image with a set of object features via object detection, considering the strong correspondences between image objects and noun entities in the source sentence.
Yang et al. \cite{yang2020visual_agreement} propose to jointly train source-to-target and target-to-source translation models to encourage the model to share the same focus on object regions by visual agreement regularization.
Yin et al. \cite{yin2020graph} propose to construct a multi-modal graph with image objects and source words according to an external visual grounding model for the alignment of multi-modal nodes.
However, the visual relevance between images and descriptions in PMT is more complex than MMT, which not only includes explicit correspondence such as products, but also implicit relevance such as shape, texture or even subjective style.

\vspace{2pt}
\noindent\textbf{E-commerce Related Tasks.}
Fashion-oriented tasks \cite{yang2020facad,zhang2020poet,zhang2020video_titling,guo2019fashion,han2017fashion_compatibility} have been recently studied due to the rapid development of e-commerce platforms.
Han et al. \cite{han2017fashion_compatibility} propose to learn compatibility relationships among fashion items for fashion recommendation.
Zhang et al. \cite{zhang2020video_titling} propose to generate a short title for user-generated videos.
Yang et al. \cite{yang2020facad} and Zhang et al. \cite{zhang2020poet} propose to generate detailed descriptions for product images/videos.
Guo et al. \cite{guo2019fashion} propose to retrieve expected products given a similar product image and a modifying text description.
However, no previous studies focus on fashion product description translation.

\vspace{2pt}
\noindent\textbf{Transformer-based Pre-training.}
Recently, self-supervised pre-training has been witnessed a great success on multiple down-stream tasks.
Devlin et al. \cite{devlin2019bert} propose a monolingual pre-trained language model named BERT and achieves the state-of-the-art results on eleven natural language processing tasks via simply pre-training with two self-supervised tasks.
Afterwards, multilingual pre-trained language models \cite{devlin2019bert, conneau2019xlm, conneau2020xlmr} and multimodal pre-trained models \cite{lu2019vilbert, li2019visualbert, chen2019uniter, li2020oscar, zhou2020unified, li2020unicoder} are further proposed to support multilingual or multimodal scenarios.
The multilingual pre-training models show strong abilities on zero-shot cross-lingual transfer, while the vision-language pre-training models outperform the state-of-the-art task-specific models on both vision-language understanding tasks, e.g., image retrieval \cite{faghri2018vsepp}, and generation tasks, e.g. image captioning \cite{vinyals2015showtell}.
However, there are few works \cite{ni2021m3p,zhou2021uc2} studying the multimodal multilingual pre-training models, and none of them verify the effectiveness on MMT or PMT tasks.
\section{Fashion-MMT Dataset}
\begin{figure}[t]
  \centering
  \begin{overpic}[scale=0.48]{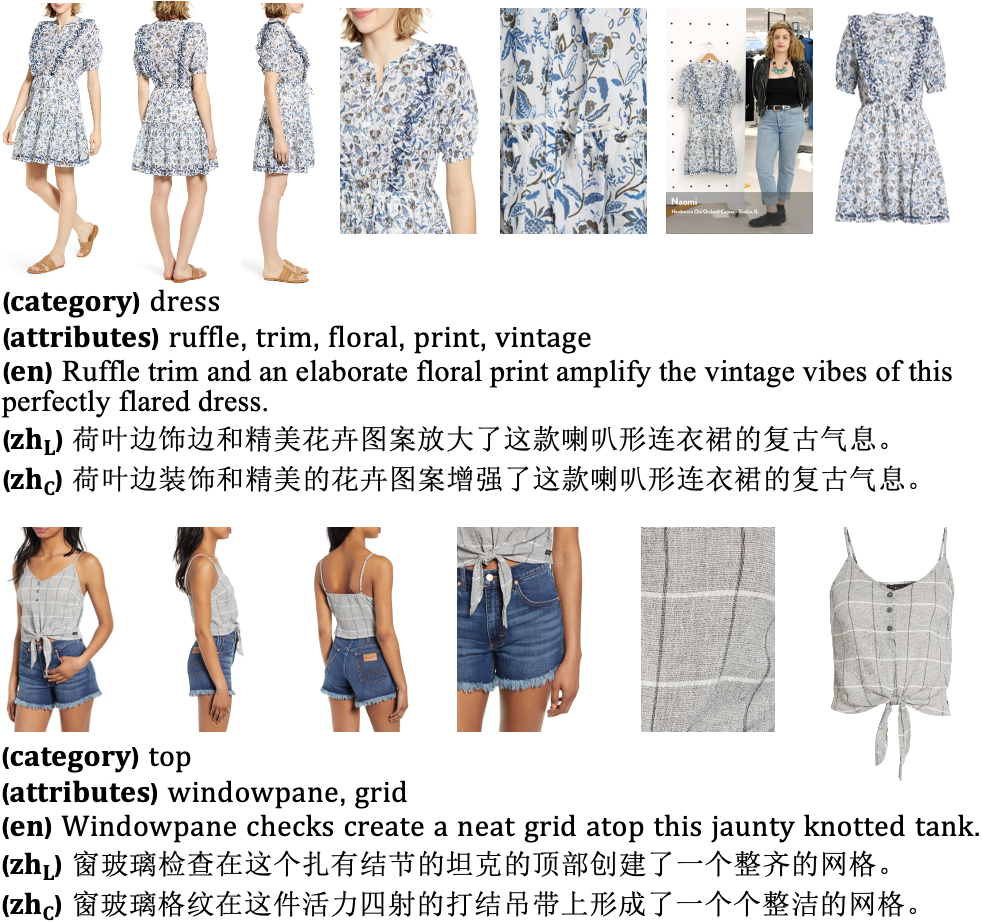}
  \end{overpic}
  %\vspace{-4pt}
  \caption{Examples from Fashion-MMT(L) and Fashion-MMT(C) datasets.}
  \label{fig:data}
\end{figure}

\vspace{2pt}
\subsection{Data Collection}
\label{sec:data_collection}
We build the Fashion-MMT dataset based on the fashion captioning dataset FACAD \cite{yang2020facad}.
The FACAD dataset contains 126,753 English product descriptions with corresponding product images crawled from the Nordstrom website. Each description is aligned with average 6\textasciitilde 7 product images of different colors and poses.
There are also product category and attributes labels for each product.

To extend the FACAD dataset for the PMT task, we first automatically translate the English product descriptions into Chinese with Google English$\rightarrow$Chinese translation system.
We remove redundant examples and the examples whose English description is shorter than 6 words, or longer than 35 words, or the Chinese translation is shorter than 15 characters.
Finally, we obtain 114,257 English-Chinese parallel description pairs with 885,244 product images, denoted as the Fashion-MMT(L) dataset.

Since the automatically translated Chinese translations contain noise, we manually clean a subset of the Fashion-MMT(L) dataset to further benefit the field.
We sample 40,000 triplets from Fashion-MMT(L) dataset and conduct manual annotation on the Alibaba Data Service platform \footnote{https://smartdata.taobao.com}.
All workers are native Chinese speakers with good English skills, and are asked to modify the automatically translated Chinese description given the images and English description.
To ensure annotation quality, each translation is further reviewed and accepted by another two independent workers.
The whole data collection window is around one month.
More than 98.5\% of automatic Chinese translations have been post-edited, and the average Levenshtein distance between before and after manual edition for each sentence is 20.2. It suggests that the state-of-the-art text-based machine translation system is far from perfect to translate product descriptions.
We denote the cleaned version as Fashion-MMT(C) dataset.
Examples from Fashion-MMT(L) and Fashion-MMT(C) datasets are presented in Figure~\ref{fig:data}.

\begin{table}
\centering
\caption{Data splits of Fashion-MMT datasets.}
%\vspace{-5pt}
\label{tab:data_split}
\small
\begin{tabular}{l| c| c c c}
\toprule
Dataset & Split & Train & Validation & Test \\
\midrule
\multirow{2}{*}[-0.5ex]{Fashion-MMT(L)} & \#Triplets & 110,257 & 2,000 & 2,000 \\
& \#Images & 853,503 & 15,974 & 15,767 \\
\midrule
\multirow{2}{*}[-0.5ex]{Fashion-MMT(C)} & \#Triplets & 36,000 & 2,000 & 2,000 \\
& \#Images & 280,915 & 15,974 & 15,767 \\
\bottomrule
\end{tabular}
\end{table}

\begin{table}
\centering
\caption{Comparison with other MMT datasets. Avg\_len denotes the average English sentence length. Avg\_img denotes the average number of images for each parallel text pair.}
%\vspace{-5pt}
\label{tab:data_compare}
\small
\begin{tabular}{l|cccc}
\toprule
Dataset & \#Image & \#Triplet & Avg\_len & Avg\_img \\
\midrule
Multi30k \cite{elliott2016Multi30k} & 31,014 & 31,014 & 13.1 & 1.00 \\
IKEA \cite{zhou2018ikea} & 3,600 & 3,600 & 71.4 & 1.00 \\
\midrule
Fashion-MMT(C) & 312,656 & 40,000 & 20.8 & 7.82 \\
Fashion-MMT(L) & 885,244 & 114,257 & 22.6 & 7.75 \\
\bottomrule
\end{tabular}
\end{table}

\subsection{Dataset Statistics}
We split each version of Fashion-MMT into three sets as shown in Table~\ref{tab:data_split}.
Since the validation and test set should be clean to correctly evaluate different models, for the Fashion-MMT(L) dataset, we use the same clean validation and test set with the Fashion-MMT(C) dataset and exclude the corresponding 4,000 triplets from the training set.

In Table~\ref{tab:data_compare}, we compare our Fashion-MMT datasets with other existing MMT datasets.
Both of our Fashion-MMT(C) and Fashion-MMT(L) datasets have more triplets than the Multi30k \cite{elliott2016Multi30k} benchmark and the commercial domain IKEA \cite{zhou2018ikea} dataset.
Compared with the IKEA dataset, our Fashion-MMT datasets covers more diverse fashion product categories, while the IKEA dataset only focuses on furnishings.
Furthermore, for each parallel text pair in the Fashion-MMT, there are multiple images associated with it, while previous MMT datasets only contain one image for each parallel text.
Multiple product images are more common in the real-world scenario as products on the e-commerce platform are usually associated with more than one image with different poses or colors, showing various product details.

\begin{figure*}[t]
  \centering
  \begin{overpic}[scale=0.53]{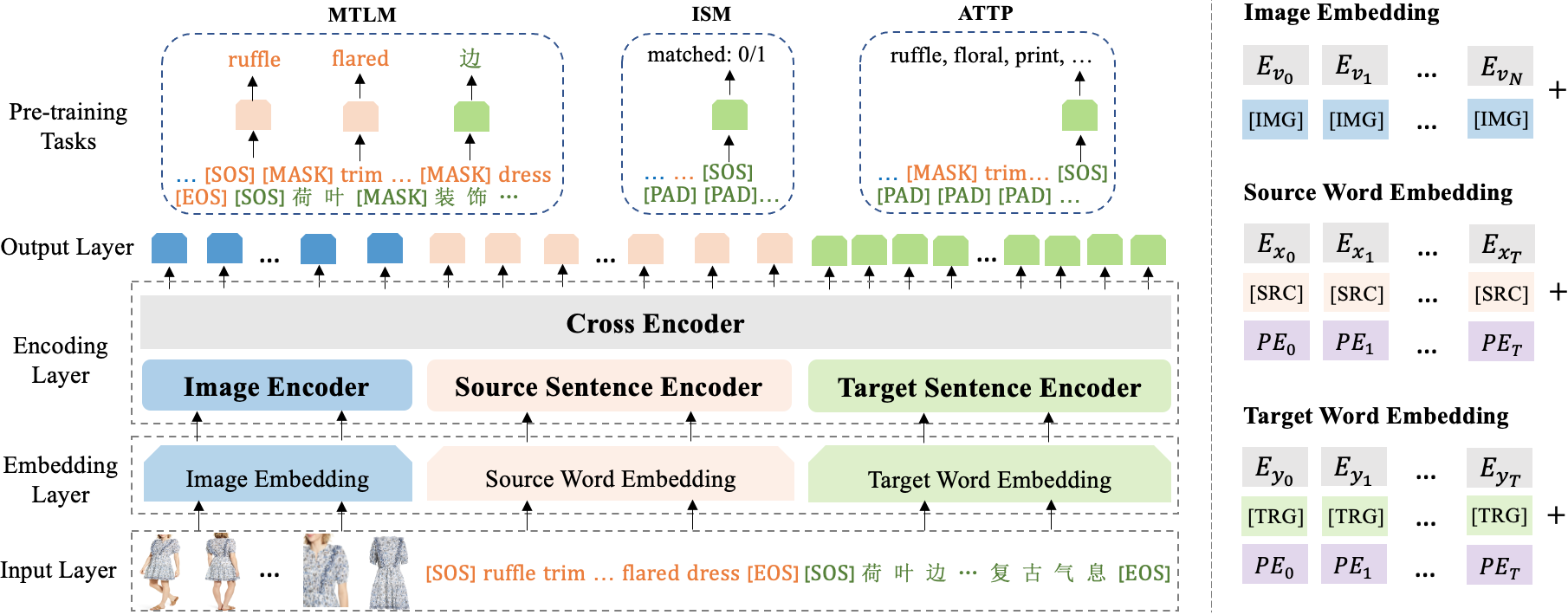}
  \end{overpic}
  %\vspace{-3pt}
  \caption{Illustration of the proposed \upoc~ model based on the cross-modal cross-lingual pre-training framework. Three independent transformer encoders are first employed to capture the intra context of each modality and language, and then a multi-layer cross encoder is adopted to encode the multimodal and multilingual context via three pre-training tasks.}
  \label{fig:model}
  %\vspace{-5pt}
\end{figure*}

\vspace{5pt}
\section{$\text{UPOC}^\text{2}$: A Unified Pre-training and Fine-tuning Framework for PMT}
In this section, we introduce our \upoc~~model and the proposed cross-modal cross-lingual pre-training tasks for the product description translation.
Figure~\ref{fig:model} illustrates the architecture of our \upoc~~model.
It is stacked with multiple multi-layer transformer encoders and follows a unified pre-training and fine-tuning scheme \cite{devlin2019bert,zhou2020unified,li2020oscar}.
We first pre-train our model with three cross-modal cross-lingual pre-training tasks to effectively learn semantic alignments between images and bilingual texts, including multimodal translation language modeling (MTLM), image source-sentence matching (ISM) and product attribute prediction (ATTP).
Then, we fine-tune the pre-trained model for the PMT task.

\subsection{Model Architecture}
\noindent\textbf{Input Representation.}
The input sequence is the concatenation of Image-Source-Target triplet ($V$, $X$, $Y$), where $V$ is the global embedding sequence of product images, $X$ is the word embedding sequence of the source language product description and $Y$ is the word embedding sequence of the target language translation.
For each $v_i$ in $V=\{v_0,\cdots,v_i,\cdots,v_N\}$, we extract the global visual feature for the $i$-th image via ResNet-101 \cite{he2016resnet}.
Then, a linear layer is employed to map the visual feature to the same dimensionality as the word embedding.
For the source and target sentences, we add a special start token ([SOS]) and a special end token ([EOS]) to the start and end of each sentence, and represent each token in the sentence with a word embedding vector learned from scratch.
To distinguish different modalities and languages of each element in the whole input sequence ($V$, $X$, $Y$), we add a learnable modality embedding
to each element indicating whether it belongs to the image modality, source language sentence or the target language sentence.
Specially, we also add positional embeddings to the tokens of source and target language sentences.
The illustration of input representations is shown in the right part of Figure~\ref{fig:model}.

\vspace{2pt}
\noindent\textbf{Multimodal Transformer.}
Our \upoc~~model contains four multi-layer transformer \cite{vaswani:transformer} encoders, including three independent encoders and a cross encoder.
We first employ independent encoders to encode the image sequence, source sentence and target sentence to capture their intra-context information, which have $L_v$, $L_s$ and $L_t$ layers respectively.
The outputs of the three encoders are then concatenated as a whole input sequence to the cross encoder.
The cross encoder is also a multi-layer transformer encoder, which is used to encode the inter-context across different modalities and languages.
We denote the number of cross encoder layers as $L_c$, hidden size as $H$, and the number of self-attention heads as $A$.
We share the parameters of the source sentence encoder and the target sentence encoder in pre-training, and separate them when adapting to the PMT task. Because the target sentence encoder will be learned as the translation generator in the fine-tuning stage.

\subsection{Pre-training Tasks}
To learn the cross-lingual correspondence between source and target language sentences and enhance the cross-modal fusion between images and texts for better translation, we pre-train our model with three pre-training tasks, described in this section.

\noindent\textbf{Task \#1: Multimodal Translation Language Modeling (MTLM)}
The cross-lingual alignment is important to machine translation.
Inspired by the multilingual pre-training task (translation language modeling (TLM) proposed in XLM \cite{conneau2019xlm}), and the multimodal pre-training task (masked language modeling (MLM) generally used in V+L pre-training models \cite{chen2019uniter,li2020unicoder,lu2019vilbert}), we propose to combine them for the multimodal multilingual scenario as the multimodal translation language modeling (MTLM) task for the PMT.

The goal of MTLM is to predict the masked words in both languages with the context information of images, the surrounding words in the same language, and all the words in the other language.
We randomly choose 15\% word tokens in both languages for prediction.
Each chosen word is replaced with a special [MASK] token 80\% of the time, another random word 10\% of the time and the original word 10\% of the time.
Note that the random words could be foreign words.
The MTLM task takes the fused feature from cross encoder as the input and predict the original word with an output softmax layer, which is tied with input embeddings. 
We share the vocabulary and softmax prediction layer across languages.
The training objective of MTLM can be expressed as follows:
\begin{equation}
    \mathcal{L}_{MTLM} = - \mathbb{E}_{(V,X,Y)\sim\mathcal{D}} \log p(x_m,y_m|x_{\setminus m},y_{\setminus m},V;\Theta)
\end{equation}
where $\mathcal{D}$ denotes the whole training set, $x_m$ and $y_m$ denote the masked words in $X$ and $Y$, and $\Theta$ denotes all learnable parameters of the pre-training model.
Note that $x_m$ and $y_m$ are not semantically aligned words.
With the MTLM pre-training task, the model learns the semantic alignments between source and target language words.
However, since the translation guidance from the images is much weaker than that from the source words, the model tends to ignore the visual modality with only the MTLM task for the target translation.
Therefore, we further propose to enhance the cross-modal fusion between images and texts via cross-modal pre-training tasks.

\vspace{2pt}
\noindent\textbf{Task \#2: Image Source-Sentence Matching (ISM)}
The cross-modal matching task has been widely used in vision-language pre-training models, which is helpful to learn the semantic alignment between the visual and textual modalities.
Considering that for the PMT/MMT task, the effective fusion between image and source sentence is important, we conduct the semantic matching between the image and source sentence.
We pad the target language sentence $Y$ except for the start [SOS] token, which is further used to predict whether the images $V$ and source sentence $X$ are semantically matched.
Specifically, the output of the [SOS] token is fed to a linear layer with the sigmoid function to predict a matching score $s(V, X)$ between 0 and 1.
We construct the negative pair by replacing the source sentence in a matched pair with another one.
Since different types of products are significantly different, to avoid an oversimplified ISM task, we choose hard negatives by randomly sampling negative source sentences from the set which describes products in the same category as the original sentence. 
In this way, the model will focus more on the product details rather than the product category to determine whether the image and description are matched.
The training objective of the ISM task can be expressed as follows:
\begin{equation}
    \mathcal{L}_{ISM} = - \mathbb{E}_{(V,X)\sim\mathcal{D}}[l \log s(V, X; \Theta) + (1-l) \log (1-s(V, X; \Theta))]
\end{equation}
where $l\in[0,1]$ indicates whether the input image-source pair is a negative or positive sample.
With the ISM pre-training task, the start [SOS] token of the target language sentence will encode rich fused information from images and the source language sentence to benefit the target language translation.

\vspace{2pt}
\noindent\textbf{Task \#3: Product Attribute Prediction (ATTP)}
The product attributes describe important information of the commercial products, including the decorations, shapes, colors or styles of the product as shown in Figure~\ref{fig:data}.
To help the model to extract these information from images for better translation, we propose to further enhance the semantic encoding of the [SOS] token by predicting the product attributes according to the images.
Specifically, we extend the binary classification of the ISM task to a multi-class classification task for the attribute prediction.
We employ another linear layer with softmax function to predict the product attributes based on the output of [SOS] token.
Since the ground-truth attributes in FACAD dataset \cite{yang2020facad} is extracted from the source sentence (the nouns and adjectives in the source sentence), it is easy to predict attributes with the source sentence as context.
Therefore, for the ATTP task, we mask the source words that express product attributes, and force the model to predict attributes relying on the images.
We adopt the categorical cross-entropy as the training objective as follows:
\begin{equation}
    \mathcal{L}_{ATTP} = - \mathbb{E}_{(V,X)\sim\mathcal{D}} \sum_{c\in C} \log p_c(V,X;\Theta)
\end{equation}
where $C$ denotes the ground-truth attributes set.

\subsection{Fine-tuning in PMT task}
After pre-training with the above three pre-training tasks, we adapt our pre-trained model to the PMT task.
The PMT/MMT task generates the target translation with the context of source sentence and images.
It is a generation task, hence, the current generated word cannot ``see'' the future words.
Therefore, we adapt our bi-directional pre-trained model to a uni-directional generator by constraining the self-attention mask of the target language sentence.
Similar to the MTLM pre-training task, we randomly choose 15\% of target sentence words and replace them with the [MASK] token 80\% of the time, another random word 10\% of the time and the original word 10\% of the time.
We predict the masked target word with the context information from all the images, all the source sentence words and the target sentence words before its position.
The training objective can be expressed as follows:
\begin{equation}
    \mathcal{L}_{PMT} = - \mathbb{E}_{(V,X,Y)\sim\mathcal{D}} \log p(y_m|y_{<m},X,V;\Theta)
\end{equation}
where $y_m$ denote the masked words in $Y$, and $\Theta$ denotes all learnable parameters of the pre-trained model.

During inference time, we first encode the images, source sentence and the start [SOS] token of the target translation as the input.
The [SOS] token encodes rich fused information from source sentence and images due to the ISM and ATTP pre-training tasks.
Then, we start to generate the target language translation word by word through feeding a [MASK] token and sampling the predicted token from the softmax output layer.
At the next step, the previous [MASK] token is replaced by the generated token, and a new [MASK] token is fed to the model for the next word generation until a [EOS] token is generated.
\section{Experiments}
We evaluate our \upoc~~model on the Multi30k \cite{elliott2016Multi30k} benchmark dataset and the new proposed Fashion-MMT(C/L) datasets.
To demonstrate the effectiveness of our cross-modal cross-lingual pre-training scheme for the PMT/MMT task, we experiment with two settings: 1) pre-training on the same downstream dataset without extra data and 2) pre-training with additional noisy triplet data.

\begin{table*}[ht]
\caption{PMT results with different pre-training tasks and pre-training data on Fashion-MMT(C) dataset.}
\vspace{-5pt}
\label{tab:aba_task}
\centering
\small
\begin{tabular}{c |c |c c c| c c c|c c c}
\toprule
\multirow{2}{*}[-0.5ex]{Row} & \multirow{2}{*}[-0.5ex]{Extra data} & \multicolumn{3}{c|}{Pre-training tasks} & \multicolumn{3}{c|}{Validation} & \multicolumn{3}{c}{Test} \\ 
\cmidrule{3-11}
& & MTLM & ISM & ATTP & BLEU@4 & METEOR & CIDEr & BLEU@4 & METEOR & CIDEr \\
\midrule
1 & - & & & & 39.39 & 34.69 & 285.42 & 39.41 & 34.59 & 281.94 \\
\specialrule{0.05em}{1.5pt}{1.5pt}
2 & - & \checkmark & & & 41.45 & 35.79 & 307.61 & 41.17 & 35.55 & 303.58 \\
3 & - & \checkmark & \checkmark & & 41.79 & 35.84 & 308.59 & 41.38 & 35.68 & 305.21 \\
4 & - & \checkmark & \checkmark & \checkmark & \textbf{41.93} & \textbf{36.01} & \textbf{313.46} & \textbf{41.56} & \textbf{35.87} & \textbf{306.82} \\
\specialrule{0.05em}{1.5pt}{1.5pt}
5 & Fashion-MMT(L) & \checkmark & \checkmark & \checkmark & \textbf{43.06} & \textbf{36.64} & \textbf{324.36} & \textbf{43.00} & \textbf{36.68} & \textbf{324.46} \\
\bottomrule
\end{tabular}
%\vspace{-5pt}
\end{table*}

\subsection{Experimental Settings}
\noindent\textbf{Datasets.}
Multi30k \cite{elliott2016Multi30k} is the benchmark dataset for the conventional MMT task, where the images are about human daily activities.
It contains 29,000 images for training, 1,014 for validation and 1,000 for testing (aka Test2016).
Each image is annotated with an English description and its human translation to German (DE).
Besides the official testing set of Multi30k, we also evaluate on another testing set (aka Test2017) released for the multimodal translation shared task \cite{elliott:ambcoco}, which is more difficult to translate.
We adopt the byte pair encoding \cite{sennrich:bpe} with 10,000 merge operations to get both the source and target language vocabularies.
For the Fashion-MMT(L/C), we segment the Chinese translation with raw Chinese characters to get the target language vocabulary.

\noindent\textbf{Metrics.}
We follow previous works \cite{Ive2019distilling, yin2020graph} to use BLEU \cite{papineni:bleu} and METEOR \cite{michael:meteor} to evaluate the translation qualities.
The BLEU metric computes the precision of $n$-grams and the METEOR metric employs more flexible matching, including exact, stem, synonym, and paraphrase matches between words and phrases.
Furthermore, we also report the CIDEr \cite{vedantam2015cider} score, which is commonly used in the image/video captioning tasks and pays more attention to visually relevant words based on the term frequency and inverse document frequency (tf-idf).

\noindent\textbf{Implementation Details.}
We pre-train two model variants for different experimental settings.
For the setting with no extra pre-training data, we set the layer number of independent encoders $L_v=L_s=L_t=1$, the layer number of cross encoder $L_c=3$, the hidden size $H=512$, and the head number $A=8$.
For the setting of data augmentation with noisy data, we drop the independent encoders to avoid the influence of noise from target translations, and set the the layer number of cross encoder $L_c=6$.
An ablation study of the layer number settings can be found in Section~\ref{sec:ablation}.
We initialize all the model parameters from scratch.

When pre-training the model, we sample the batch of each pre-trained task with the proportion of MTLM:ISM:ATTP=9:2:1 except for Multi30k.
Since there is no attribute in the Multi30k dataset, we pre-train the model without the ATTP task, and sample the batch of the two other tasks with the the proportion of MTLM:ISM=3:1.
We pre-train our model with at most 250K iterations and a warm-up strategy of 10,000 steps.
When fine-tuning the pre-trained model on PMT/MMT task, we set the learning rate as 6e-5.

For the images in Fashion-MMT(C/L), we use the ResNet-101 \cite{he2016resnet} pre-trained on ImageNet to encode the global feature for each image.
To narrow the large domain gap between ImageNet and our Fashion-MMT dataset, we fine-tune the conv4\_x and the conv5\_x layers of ResNet-101 with the objectives of product category classification and attribute prediction.
For the images in Multi30k, we follow previous works \cite{yin2020graph,lin2020capsule} to represent each image with object region features.
We detect up to 20 object regions and extract their corresponding visual features by Faster R-CNN \cite{ren2015fastrcnn}.

\begin{figure}[t]
  \centering
  \begin{overpic}[scale=0.27]{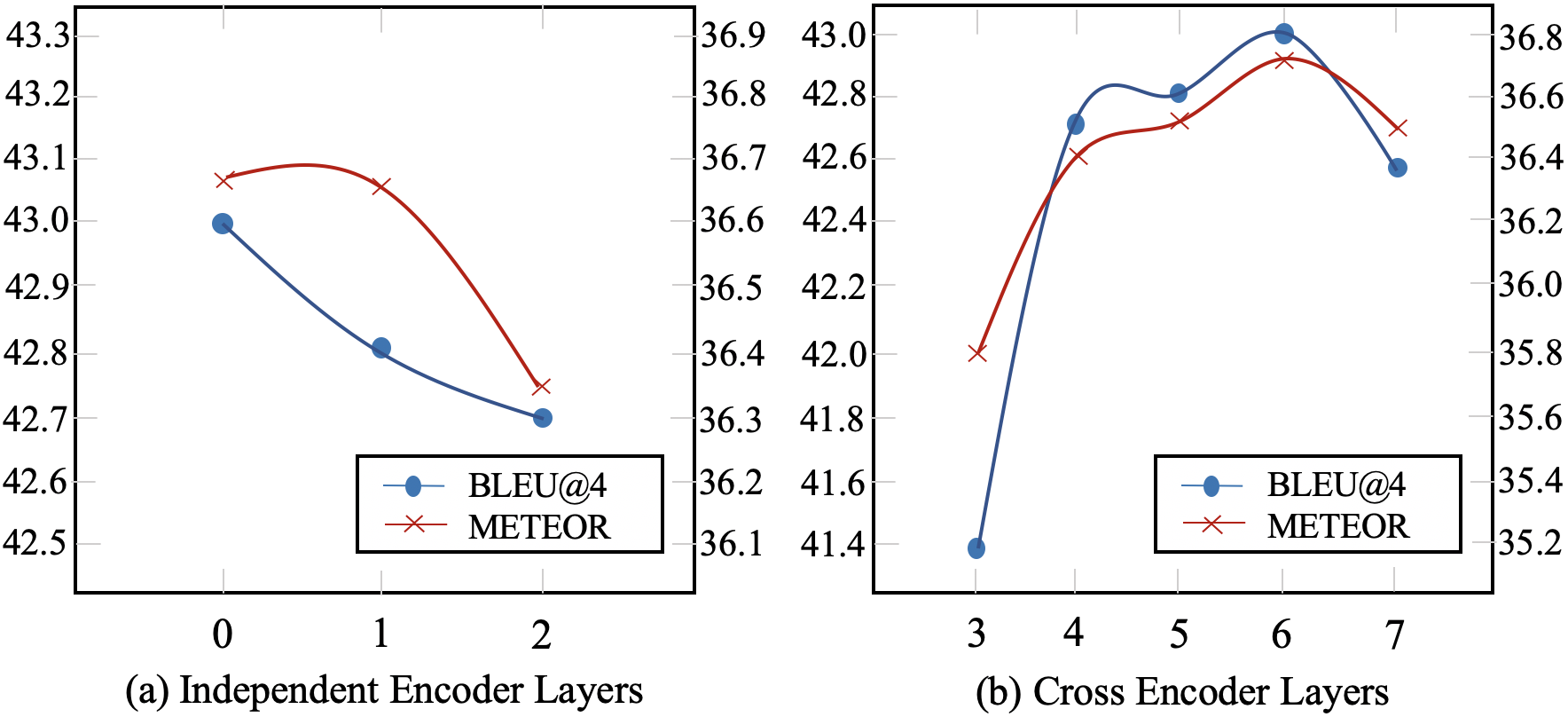}
  \end{overpic}
  \vspace{-5pt}
  \caption{Results variation with different layer numbers of encoders on Fashion-MMT test set.}
  \vspace{-10pt}
  \label{fig:aba_layer}
\end{figure}

\begin{table*}[ht]
\caption{Experimental results on EN$\Rightarrow$ZH translation task on the Fashion-MMT(C) dataset.}
\vspace{-4pt}
\label{tab:FAMMT_results}
\small
\centering
\begin{tabular}{c |l| c| c c c|c c c}
\toprule
\multirow{2}{*}[-0.5ex]{Row} & \multirow{2}{*}[-0.5ex]{Method} & \multirow{2}{*}[-0.5ex]{Extra data} & \multicolumn{3}{c|}{Validation} & \multicolumn{3}{c}{Test} \\ 
\cmidrule{4-9}
& & & BLEU@4 & METEOR & CIDEr & BLEU@4 & METEOR & CIDEr \\
\midrule
1 & Transformer \cite{vaswani:transformer} & - & 40.58 & 35.84 & 303.69 & 40.61 & 35.77 & 302.3 \\
2 & Multimodal Graph \cite{yin2020graph} & - & 41.07 & 35.55 & 307.38 & 40.70 & 35.45 & 305.08 \\
3 & \textbf{$\text{UPOC}^\text{2}$ (ours)} & - & \textbf{41.93} & \textbf{36.01} & \textbf{313.46} & \textbf{41.56} & \textbf{35.87} & \textbf{306.82} \\
\midrule
4 & Transformer \cite{vaswani:transformer} & Fashion-MMT(L) & 41.28 & 36.06 & 315.15 & 41.21 & 35.91 & 312.34 \\
5 & Multimodal Graph \cite{yin2020graph} & Fashion-MMT(L) & 41.39 & 35.96 & 316.21 & 41.49 & 35.95 & 312.68 \\
6 & \textbf{$\text{UPOC}^\text{2}$ (ours)} & Fashion-MMT(L) & \textbf{43.06} & \textbf{36.64} & \textbf{324.36} & \textbf{43.00} & \textbf{36.68} & \textbf{324.46} \\
\bottomrule
\end{tabular}
\vspace{-4pt}
\end{table*}

\begin{table}[t]
\caption{EN$\Rightarrow$DE translation results on Multi30k dataset, where B denotes BLEU, and M denotes METEOR.}
\vspace{-4pt}
\label{tab:multi30k_results}
\small
\centering
\begin{tabular}{l |c c| c c }
\toprule
\multirow{2}{*}[-0.5ex]{Model} & \multicolumn{2}{c|}{Test2016} & \multicolumn{2}{c}{Test2017} \\
\cmidrule{2-5}
& B@4 & M & B@4 & M \\
\midrule
INIT \cite{calixto-liu:Init} & 37.3 & 55.1 & - & - \\
DATTN \cite{calixto:doublyAtt} & 36.5 & 55.0 & - & - \\
SATTN \cite{delbrouck:gating} & 38.2 & 55.4 & - & - \\
Imagination \cite{elliott:Imagination} & 36.8 & 55.8 & - & - \\
$\mathrm{\text{VMMT}_\text{F}}$ \cite{calixto:latent} & 37.7 & 56.0 & 30.1 & 49.9 \\
Deliberation \cite{Ive2019distilling} & 38.0 & 55.6 & - & - \\
VGR \cite{yang2020visual_agreement} & - & - & 29.5 & 50.3 \\
UVR \cite{zhang:universal} & 36.9 & - & 28.6 & - \\
Multimodal Transformer \cite{yao2020mmt} & 38.7 & 55.7 & - \\
Multimodal Graph \cite{yin2020graph} & 39.8 & 57.6 & 32.2 & 51.9 \\
DCCN \cite{lin2020capsule} & 39.7 & 56.8 & 31.0 & 49.9 \\
\midrule
\textbf{$\text{UPOC}^\text{2}$ (ours)} & \textbf{40.8} & \textbf{58.9} & \textbf{34.1} & \textbf{53.4} \\
\bottomrule
\end{tabular}
\vspace{-6pt}
\end{table}

\begin{table*}[t]
\caption{EN$\Rightarrow$ZH translation results on Fashion-MMT dataset with different number of clean triplets for fine-tuning.}
%\vspace{-5pt}
\label{tab:few_shot}
\small
\centering
\begin{tabular}{c| l| c| c c c|c c c}
\toprule
\multirow{2}{*}[-0.5ex]{Row} & \multirow{2}{*}[-0.5ex]{Method} & \multirow{2}{*}[-0.5ex]{\begin{tabular}[c]{@{}c@{}}Fine-tuning \\ data (\#Triplets)\end{tabular}} & \multicolumn{3}{c|}{Validation} & \multicolumn{3}{c}{Test} \\ 
\cmidrule{4-9}
& & & BLEU@4 & METEOR & CIDEr & BLEU@4 & METEOR & CIDEr \\
\midrule
1 & $\text{UPOC}^\text{2}$ & 5,000 & 38.33 & 34.06 & 281.18 & 38.13 & 33.92 & 276.74 \\
2 & $\text{UPOC}^\text{2}$ & 10,000 & 40.32 & 34.92 & 298.31 & 39.95 & 35.01 & 293.69\\
3 & $\text{UPOC}^\text{2}$ & 15,000 & 41.28 & 35.63 & 307.15 & 40.89 & 35.48 & 306.28\\
\midrule
4 & $\text{UPOC}^\text{2}$ & 36,000 (full) & 42.44 & 36.21 & 323.22 & 42.43 & 36.25 & 320.25 \\
5 & Multimodal Graph \cite{yin2020graph} & 36,000 (full) & 41.07 & 35.55 & 307.38 & 40.70 & 35.45 & 305.08 \\
\bottomrule
\end{tabular}
\end{table*}

\subsection{Pre-training for PMT}
\label{sec:ablation}
\noindent\textbf{Pre-training Tasks and Data.}
In Table~\ref{tab:aba_task}, we experiment with different pre-training tasks and data to evaluate the effectiveness of our \upoc~~model for the PMT task.
The row~1 stands for a non-pretrain baseline, which is directly trained with the fine-tuning task on Fashion-MMT(C) dataset.
When first pre-trained with the MTLM task on the same dataset, and then fine-tuned for the PMT evaluation, the model is improved significantly by over 2 points on BLEU@4 as shown in row~2, which demonstrates the effectiveness of our pre-training and fine-tuning schemes on the PMT task.
Gradually combining the ISM and ATTP tasks with the MTLM task improves the BLEU@4 score from 41.45 to 41.93, and the CIDEr score from 307.61 to 313.46, which indicates the effectiveness of ISM and ATTP tasks.
Pre-training with the three tasks together finally achieves the best translation results on the Fashion-MMT(C) dataset.
In row~5, we add the noisy triplet data from Fashion-MMT(L) dataset in the pre-training stage, and fine-tune the model on the clean Fashion-MMT(C) dataset, which achieves significant additional gains compared with row~4.
It shows that our model can benefit from machine generated noisy data which are easy to acquire by simply translating the existing extensive mono-lingual product descriptions through machine translation system.
We also evaluate our \upoc~~model in a lower-resource setting in Section~\ref{sec:few_shot}, where the model is pre-trained on the noisy Fashion-MMT(L) dataset and fine-tuned with less clean PMT data.

\vspace{3pt}
\noindent\textbf{Different Layer Numbers of Encoders.}
In our \upoc~~model, there are three independent encoders to encode the intra-context information for each modality and language, and a cross encoder to capture the cross-modal and cross-lingual information.
We set the layer number of independent encoders as 1 and that of the cross encoder as 3 when only using clean data in the pre-training.
However, when adding more noisy data for pre-training, we study the translation performance with respect to different layer numbers in Figure~\ref{fig:aba_layer}.
The left figure shows that increasing the layer number of independent encoders will lead to performance degradation, because the noise in the machine generated target translations will influence the language modeling of target sentence encoder.
Therefore, we drop independent encoders in this setting and directly encodes the concatenated input sequence with a unified cross encoder.
In the right figure, we experiment with different cross encoder layers.
It shows that with the number of cross encoder layers increasing, the translation performance improves until the layer number exceeds 6.
It is also found that when the number of layers are less than 4, the model achieves even worse results than that without additional pre-training data, which implies that too simple model architecture will hinder the performance gains from data augmentation.

\subsection{Comparison with the State-of-the-arts}
To demonstrate the effectiveness of our proposed \upoc~~model in the PMT and MMT tasks, we compare our model with the following state-of-the-art NMT and MMT baselines:
\parskip=0.1em
\begin{itemize}[itemsep=1.5pt,partopsep=0pt,parsep=\parskip,topsep=1.5pt]
    \item Transformer \cite{vaswani:transformer}: The state-of-the-art text purely NMT model.
    \item Multimodal Graph \cite{yin2020graph}: The state-of-the-art MMT model, which constructs a multimodal graph with the source sentence and image.
\end{itemize}

Table~\ref{tab:FAMMT_results} and Table~\ref{tab:multi30k_results} report the translation results of different models on the Fashion-MMT(C) and Multi30k datasets respectively.
Even without any extra data for pre-training, our \upoc~~model outperforms the state-of-the-art MMT model on both datasets, which demonstrates the effectiveness of our pre-training and fine-tuning strategies for PMT and MMT tasks.
Comparing row~1 and row~2 in Table~\ref{tab:FAMMT_results}, it shows that the conventional state-of-the-art MMT model has limited improvements over the pure text NMT model when adapting to the PMT task.
It further validates the challenge of the PMT task, that is, the visual relevance between the product image and description is more complex than that in image captioning data and it is difficult to capture via simply connecting the source entities with visual objects.
However, our \upoc~~model achieves the best performance on both PMT and conventional MMT tasks even without any extra data for pre-training, due to its good cross-modal fusion abilities through multiple pre-training tasks.

When pre-training with more noisy triplet data annotated by machine translation system, our model achieves additional gains, while the NMT and state-of-the-art MMT models receive limited improvement due to the noise influence.
%The results in row~4 and row~5 in Table~\ref{tab:FAMMT_results} further show that the more complex models are more susceptible to data noise.
However, with the proposed multiple cross-modal cross-lingual pre-training tasks, our model has better abilities to alleviate the influence of noise from the training data than the models solely based on the maximum likelihood estimation (MLE) objective over the ground-truth translation.

\subsection{Lower-resource Learning Results}
\label{sec:few_shot}
Considering that the manually annotated PMT triplets are rather rare in reality, while there are a lot of mono-lingual product descriptions available at the e-commerce platform, we conduct experiments in a lower-resource setting where our \upoc~~model is pre-trained only with machine generated noisy triplets in Fashion-MMT(L) and fine-tuned with fewer clean triplet data in Fashion-MMT(C). 
Table~\ref{tab:few_shot} shows the translation results with different numbers of triplets used in the fine-tuning stage.
With only 15,000 annotated triplets (\textasciitilde40\% of the dataset) used for fine-tuning, our model achieves comparable results with the state-of-the-art MMT model (row~5), which is trained on the full set of Fashion-MMT(C).
It demonstrates that our model can greatly alleviate the demand of annotated triplets and effectively benefit from the noisy data which can be easily acquired from the existing large-scale mono-lingual product descriptions.
The result of our model fine-tuned on the full set in row~4 is slightly inferior than our final result in row~6 of Table~\ref{tab:FAMMT_results}. It is because our final model is pre-trained on both the Fashion-MMT(L) and Fashion-MMT(C) datasets, while in the lower-resource setting, we pre-train the model only on the noisy Fashion-MMT(L) dataset.

\begin{figure}[t]
  \centering
  \begin{overpic}[scale=0.49]{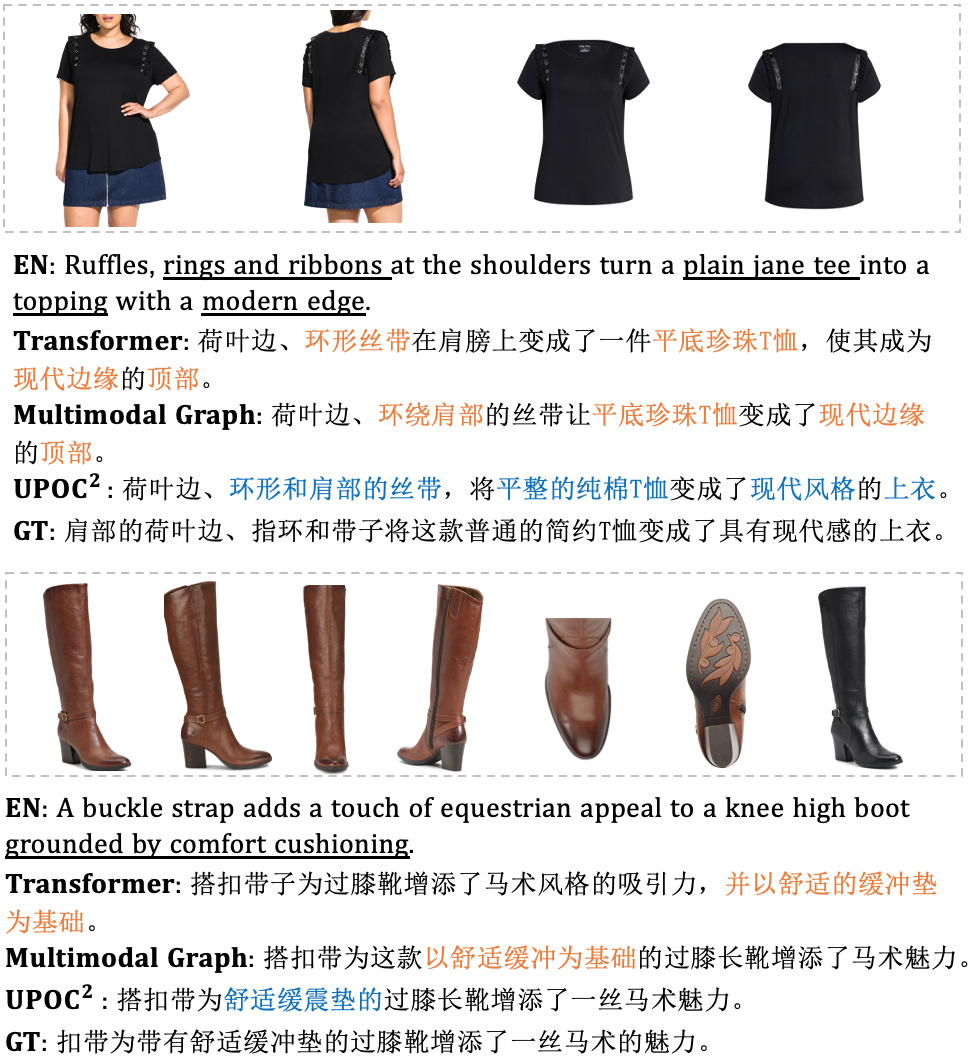}
  \end{overpic}
  %\vspace{-3pt}
  \caption{The yellow represents incorrect translations of underlined source words, while our model translates them correctly, which are colored in blue.}
  %\vspace{-10pt}
  \label{fig:results}
\end{figure}

\begin{figure}[t]
  \centering
  \begin{overpic}[scale=0.43]{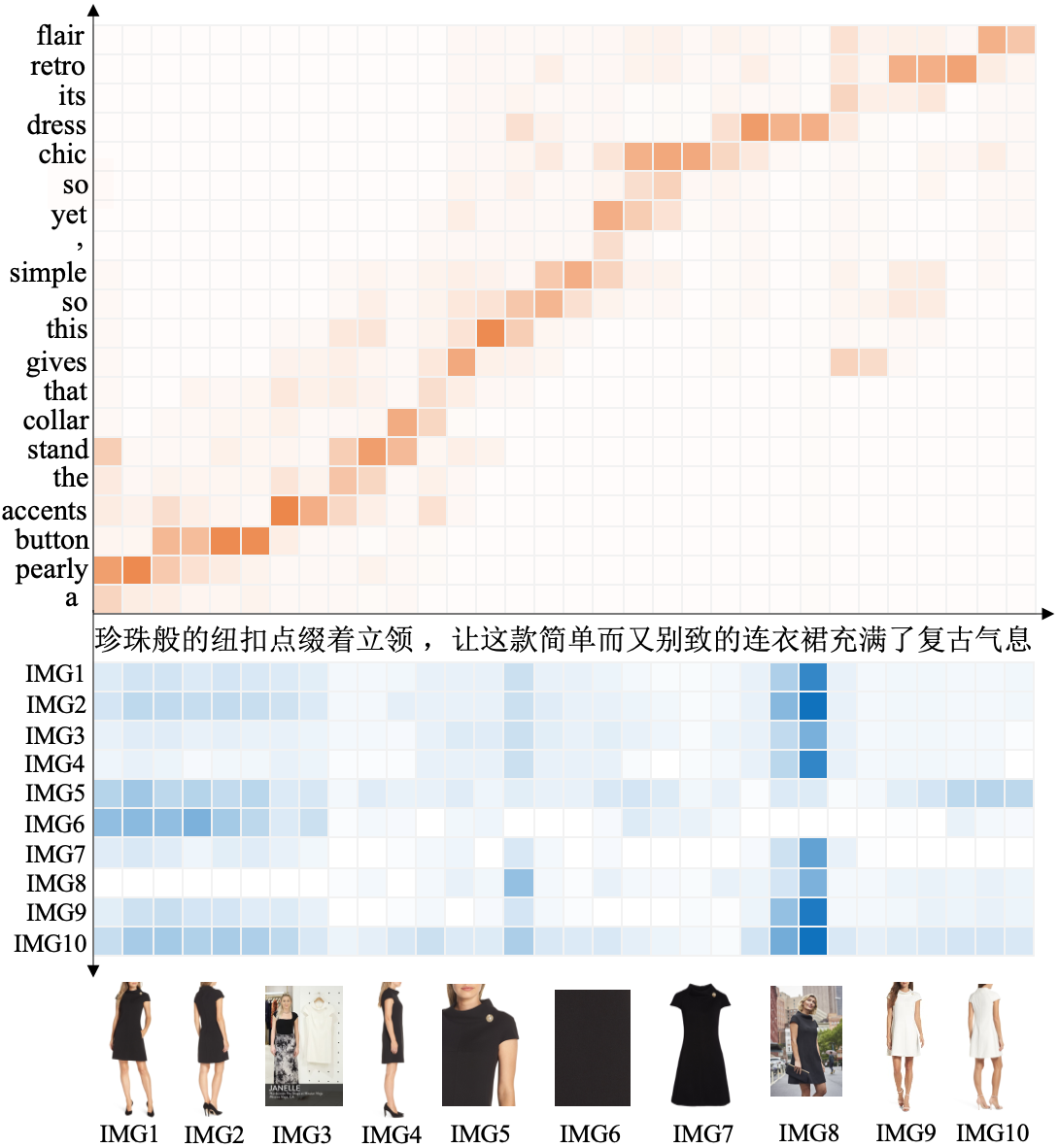}
  \end{overpic}
  %\vspace{-3pt}
  \caption{Visualization of the attention map on the images and source sentence at each EN$\Rightarrow$ZH translation step.}
  %\vspace{-10pt}
  \label{fig:attn}
\end{figure}

\subsection{Qualitative Results}
In Figure~\ref{fig:results}, we visualize the EN$\Rightarrow$ZH translation results of our \upoc~~model and other compared baselines on the Fashion-MMT(C) dataset.
Our model is pre-trained and fine-tuned on the same dataset without any extra data.
In the first example, the text-only transformer and the state-of-the-art MMT model both literally translate the source words ``edge'' and ``top'', while in this example, the word ``edge'' means ``style'' and the word ``top'' means the blouse.
With the help of the cross-modal information, our model correctly understands and translates these words.
In the second example, the NMT model and MMT model both incorrectly understand the source word ``grounded'', and translate it as the word ``foundation''.
However, our model correctly translates it with the meaning of sole.

In Figure~\ref{fig:attn}, we visualize the attention map on the source sentence and images at each translation generation step.
Each generated target word is shown to align well with the corresponding source word, which demonstrates that our model learns a good bi-lingual semantic alignment for the accurate translation.
For the visual attention, the model is shown to attend more on the detailed partial images (IMG5-6) when translating the source words ``pearly button'', while it attends more on other full body images when translating the word ``dress''.
It shows that our model can focus on the related images when translating different source words.
\vspace{3pt}
\section{Conclusion}
In this paper, we propose a large-scale product-oriented machine translation (PMT) dataset Fashion-MMT with two versions to support the PMT research.
Considering that the relevance between product images and descriptions is more complex than that in image captioning dataset, we propose a unified pre-training and fine-tuning framework called \upoc~~to enhance the cross-modal fusion ability.
We propose three cross-modal cross-lingual pre-training tasks to improve the semantic alignments between images and texts, and demonstrate its effectiveness on both the PMT and the conventional MMT tasks.
Experimental results on the proposed Fashion-MMT dataset and the Multi30k benchmark dataset show that our model outperforms the state-of-the-art models even when pre-trained on the same dataset without any extra data.
Furthermore, our model also shows the good ability to exploit more noisy triplet data to improve the translation quality and alleviate the demand of expensive manual annotations.

\begin{acks}
This work was partially supported by National Natural Science Foundation of China (No.61772535 and No.62072462), Beijing Natural Science Foundation (No.4192028), and National Key R\&D Program of China (No.2020AAA0108600).
\end{acks}

\bibliographystyle{ACM-Reference-Format}
\bibliography{references}

\end{document}